\documentclass[10pt,onecolumn,letterpaper]{article}
\usepackage{cvpr}
\usepackage[numbers,sort&compress]{natbib}
\usepackage{times}
\usepackage{eso-pic}

\usepackage{epsfig}
\usepackage{graphicx}
\usepackage{amsthm}
\usepackage{amsmath}
\usepackage{amssymb}
\usepackage[breaklinks=true,bookmarks=false]{hyperref}

\numberwithin{equation}{section}
\usepackage{datetime}

\usepackage[utf8]{inputenc} 
\usepackage[T1]{fontenc}    
\usepackage{hyperref}       
\usepackage{url}            
\usepackage{booktabs}       
\usepackage{amsfonts}       
\usepackage{nicefrac}       
\usepackage{microtype}      
\usepackage{xcolor}         

\usepackage{graphicx}
\usepackage{amsmath} 
\usepackage{algorithm}
\usepackage{algorithmic}
\usepackage{colortbl} 
\usepackage{tcolorbox}

\usepackage{multirow}
\usepackage{array}
\usepackage{longtable}
\usepackage{adjustbox}
\usepackage{array}
\usepackage{lscape}
\usepackage{natbib}
\usepackage[a4paper,margin=1in,headheight=15pt,headsep=24pt]{geometry}

\usepackage{fancyhdr}

\usepackage{amsmath,amsfonts,bm}

\def\eqref#1{equation~\ref{#1}}

\def\1{\bm{1}}

\DeclareMathAlphabet{\mathsfit}{\encodingdefault}{\sfdefault}{m}{sl}
\SetMathAlphabet{\mathsfit}{bold}{\encodingdefault}{\sfdefault}{bx}{n}

\usepackage{hyperref}
\usepackage{url}
\usepackage{graphicx}
\usepackage{multirow}
\usepackage{booktabs}
\usepackage{array}
\usepackage[table]{xcolor}
\usepackage{pifont}
\usepackage{subcaption}

\usepackage{algorithm}
\usepackage{algorithmic}
\usepackage{amsmath,bm}
\usepackage{amssymb}
\usepackage{amsthm}
\usepackage{multirow} 
\usepackage{graphicx}
\usepackage{amsmath}
\usepackage{amsthm}

\usepackage{enumitem}
\usepackage{float}
\newcommand{\downarrowtext}[1]{\textcolor{red!80!black}{\scriptsize$\downarrow$#1}}
\newcommand{\uparrowtext}[1]{\textcolor{green!60!black}{\scriptsize$\uparrow$#1}}

\usepackage{listings}
\usepackage{tcolorbox}
\tcbuselibrary{breakable}
\tcbset{
    promptbox/.style={
        breakable,
        coltitle=black,
        fonttitle=\bfseries,
        toptitle=3pt,       
        bottomtitle=1pt,    
        top=2pt,            
        bottom=4pt,         
        left=5pt,           
        right=5pt,
        arc=1pt,            
        boxrule=0.4pt,      
        colframe=gray!30,   
    }
}

\cvprfinalcopy 
\def\cvprPaperID{****} 
\allowdisplaybreaks

\begin{document}
\lhead{}
\lfoot{\date{\today},\date{\currenttime}}
\rfoot{NGD for DL}

\title{IVR-R1: Refining Trajectories through Iterative Visual-Grounded Reasoning in Reinforcement Learning}
\author{
\textbf{Chenghao Li}$^{1}$\quad
\textbf{Fusheng Hao}$^{4}$\quad
\textbf{Xikai Zhang}$^{1}$\quad
\textbf{Likang Xiao}$^{1}$\quad
\textbf{Yanwei Ren}$^{3}$\quad\\
\textbf{Fuxiang Wu}$^{4}$\quad
\textbf{Quan Chen}$^{3}$\quad
\textbf{Liu Liu}$^{2,1*}$\\[0.5em]
$^1$Hangzhou International Innovation Institute, Beihang University\\
$^2$School of Artificial Intelligence, Beihang University\\
$^3$Kuaishou Technology\\
$^4$Shenzhen Institute of Advanced Integration Technology, Shenzhen
}
\maketitle
\begingroup
\renewcommand\thefootnote{*}
\footnotetext{Corresponding author: \texttt{liuliubh@buaa.edu.cn}}
\endgroup
\begin{abstract}
  Multimodal large language models via reinforcement learning (RL) have demonstrated remarkable capabilities in complex visual reasoning tasks, yet they remain limited in long-horizon multimodal scenarios, often suffering from visual hallucination and logical error. 
  Current methods typically pre-encode high-dimensional visual scenes into discrete textual proxies to facilitate downstream reasoning. As the reasoning chain unfolds, however, the inherent information asymmetry between text and visual scenes tends to erode visual grounding, resulting in misguided reasoning and erroneous outputs.
  To address these issues, we propose IVR-R1 (Iterative Visual-grounded Reasoning), a novel RL training framework that enables dynamic visual re-alignment to rectify reasoning trajectories and improve policy optimization. To this end,we design two key components. The first is visual erosion localization, which identifies flawed rollouts through a reward-driven screening mechanism and performs fine-grained erosion localization within the multimodal context. The second is automated trajectory rectification, which iteratively cross-references intermediate reasoning states against pristine visual priors and rectifies erroneous trajectories via a Re-Reasoning Loop, thereby synthesizing expert-level demonstrations that serve as high-fidelity reasoning templates for the policy model.
  Our experiments across diverse multimodal benchmarks demonstrate that IVR-R1 consistently outperforms existing reinforcement learning methods, establishing a superior paradigm for maintaining logical and visual consistency in complex multimodal reasoning. The source code will be made publicly available.
\end{abstract}
\begin{figure}
  \centering
  \vspace{3em}
  \includegraphics[width=1\linewidth]{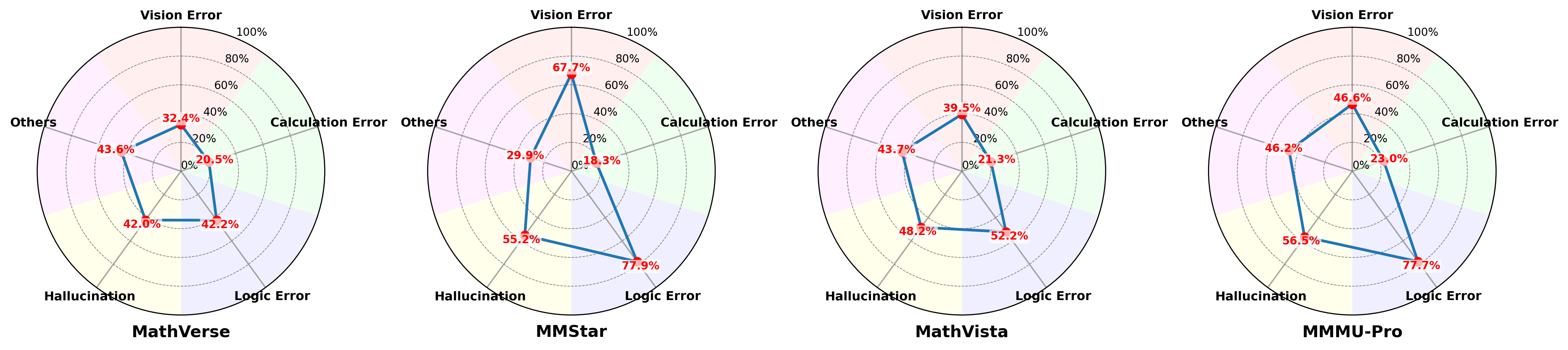}
  \caption{Illustration of the proportions of various errors for the Qwen3-vl-4b model~\cite{bai2025qwen3vltechnicalreport} across different datasets. The circular radar charts illustrate model vulnerabilities in complex visual reasoning tasks, highlighting that \textbf{high rates of logic errors and persistent hallucinations} remain the primary bottlenecks alongside significant Vision Error.}
  \label{fig:teaser_example}
  \vspace{2em}
\end{figure}
\section{Introduction}

\begin{figure*}[t]
  \centering
  \includegraphics[width=1\linewidth]{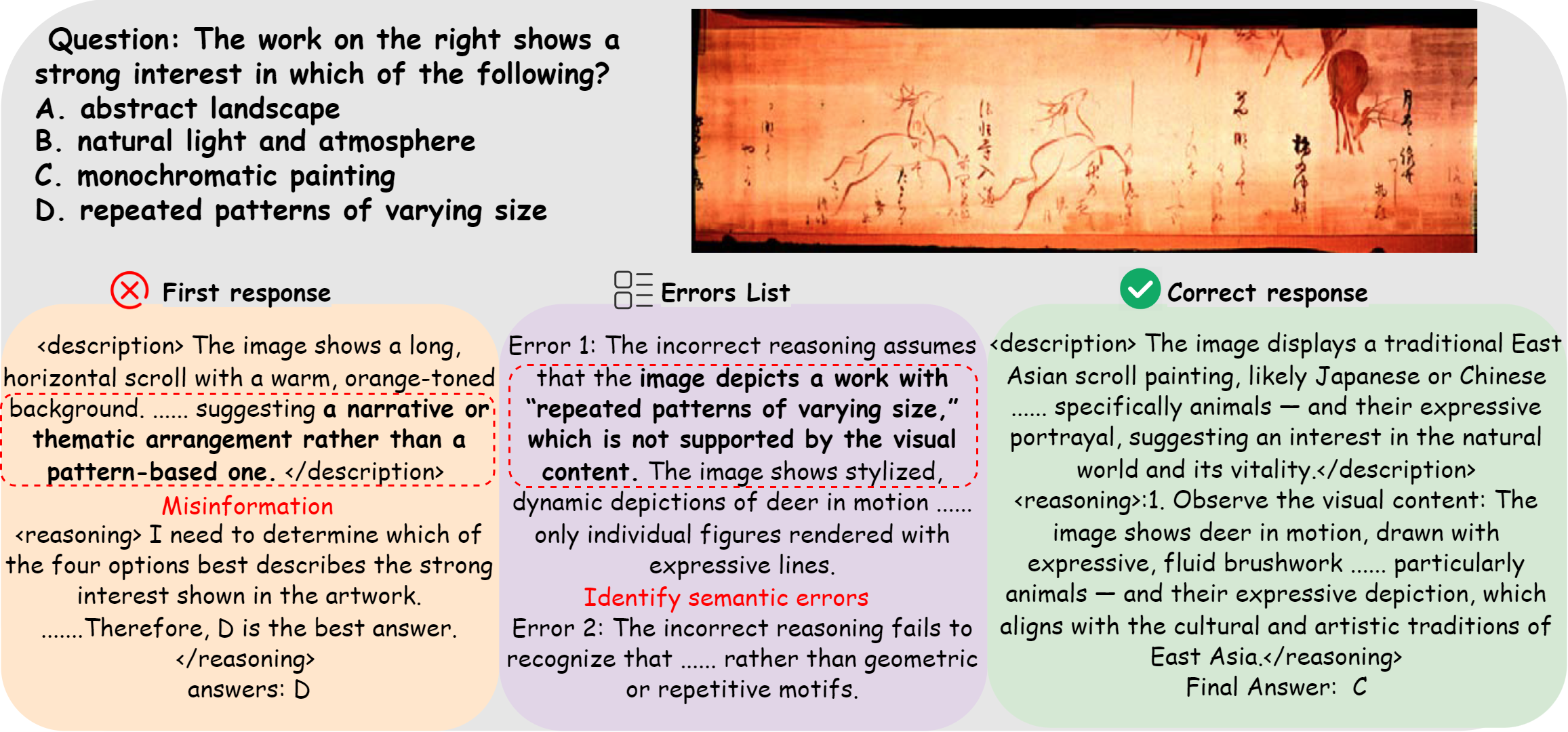}
  \caption{Illustrative example of the self-correction process in IVR-R1. (Left) The initial response fails due to visual grounding erosion, erroneously identifying "repeated patterns." (Middle) The framework performs fine-grained error attribution to identify semantic misalignments. (Right) The Re-Reasoning Loop rectifies the trajectory by re-anchoring the logic to visual evidence (e.g., fluid brushwork), leading to the correct answer.}
  \label{fig:example}
  \vspace{0.1em}
\end{figure*}

Recent advances in Multimodal Large Language Models (MLLMs) have centered on integrating pre-trained language backbones with visual encoders via instruction tuning~\cite{bai2025qwen25vltechnicalreport,bai2025qwen3vltechnicalreport,zhao2024aligngptmultimodallargelanguage,li2023blip2bootstrappinglanguageimagepretraining,lu2024deepseekvlrealworldvisionlanguageunderstanding,Yin_2024}. To further bridge the gap between perception and complex problem-solving, the field has increasingly pivoted toward post-training alignment~\cite{vanniekerk2025posttraininglargelanguagemodels,liu2024improvedbaselinesvisualinstruction,hosseini2024vstartrainingverifiersselftaught}, notably Group Relative Policy Optimization~\cite{Guo_2025}—a critic-less reinforcement learning (RL) paradigm that leverages group-relative rewards to catalyze advanced Chain-of-Thought (CoT) reasoning. While this allows models to "think" through visual evidence with greater logical depth, it often inadvertently encourages a "thinking over seeing" bias, where internal linguistic reasoning overrides external visual perception~\cite{yoon2025visualrepresentationalignmentmultimodal}. Especially in long-horizon scenarios, this imbalance leads to visual grounding erosion, causing reasoning trajectories to decouple from original visual priors~\cite{sun2025mitigatingvisualforgettingtakealong,peng2026deeperthoughtweakeraim,luo2025thinkingdriftsevidentialgrounding,sun2026thinkinghurtsmitigatingvisual}. Consequently, despite apparent improvements in logical fluency, MLLMs remains susceptible to reward hacking and persistent hallucinations, as they increasingly rely on language shortcuts from prior knowledge rather than rigorous visual anchoring~\cite{zhang2025pixelstokensbytepairencoding}.

One significant factor contributing to these failures is the information asymmetry between high-dimensional visual scenes and discrete textual proxies~\cite{sun2025mitigatingvisualforgettingtakealong,peng2026deeperthoughtweakeraim}, Figure~\ref{fig:teaser_example} quantitatively highlights the severity of this issue, revealing that a substantial and persistent Vision Error remains a critical bottleneck alongside complex logic flaws across multiple datasets. In most existing post-training methods for vision-language models, visual information is typically pre-encoded into a fixed set of tokens to facilitate downstream reasoning~\cite{li2025selfrewardingvisionlanguagemodelreasoning,xia2025visionaryr1mitigatingshortcutsvisual}. As the reasoning chain unfolds, however, the model's internal state tends to drift away from the original visual priors—a phenomenon exemplified by the failure case in Figure~\ref{fig:example}. While some methods have attempted to mitigate this modality gap by explicitly aligning internal visual representations with pre-trained vision foundation models to retain fine-grained details~\cite{yoon2025visualrepresentationalignmentmultimodal,zhang2025pixelstokensbytepairencoding}, the broader challenge of visual grounding erosion remains prevalent. This erosion leads to a progressive decoupling of reasoning logic from visual evidence~\cite{ye2024xvilacrossmodalityalignmentlarge,sun2026thinkinghurtsmitigatingvisual}, causing the model to "hallucinate" details absent from the image or follow misguided reasoning paths that contradict the visual context~\cite{sun2025mitigatingvisualforgettingtakealong,peng2026deeperthoughtweakeraim}. Consequently, even with advanced CoT capabilities, MLLMs often produce erroneous outputs when their linguistic "thinking" is no longer strictly anchored to the visual scene.

To address these challenges, we introduce IVR-R1 (Iterative Visual-grounded Reasoning), a novel RL training framework characterized by its modularized pipeline of Visual Erosion Localization and Automated Trajectory Rectification. Unlike standard RL approaches that merely penalize incorrect final answers—often resulting in a sparse and suboptimal learning signal, IVR-R1 actively rectifies the flawed rollouts by probing the internal reasoning process via two cohesive core stages. First, it employs a reward-driven screening mechanism to identify "grounding-eroded" rollouts and pinpoints fine-grained, step-level misalignments by cross-referencing intermediate logical states against the original visual features, thus enabling visual erosion localization. Second, guided by these diagnostic insights, it revises flawed reasoning paths into high-fidelity, expert-level demonstrations by re-anchoring the logic to the visual evidence, thereby realizing automated trajectory rectification.
\begin{figure*}[t] 
  \centering         
  \includegraphics[width=1\linewidth]{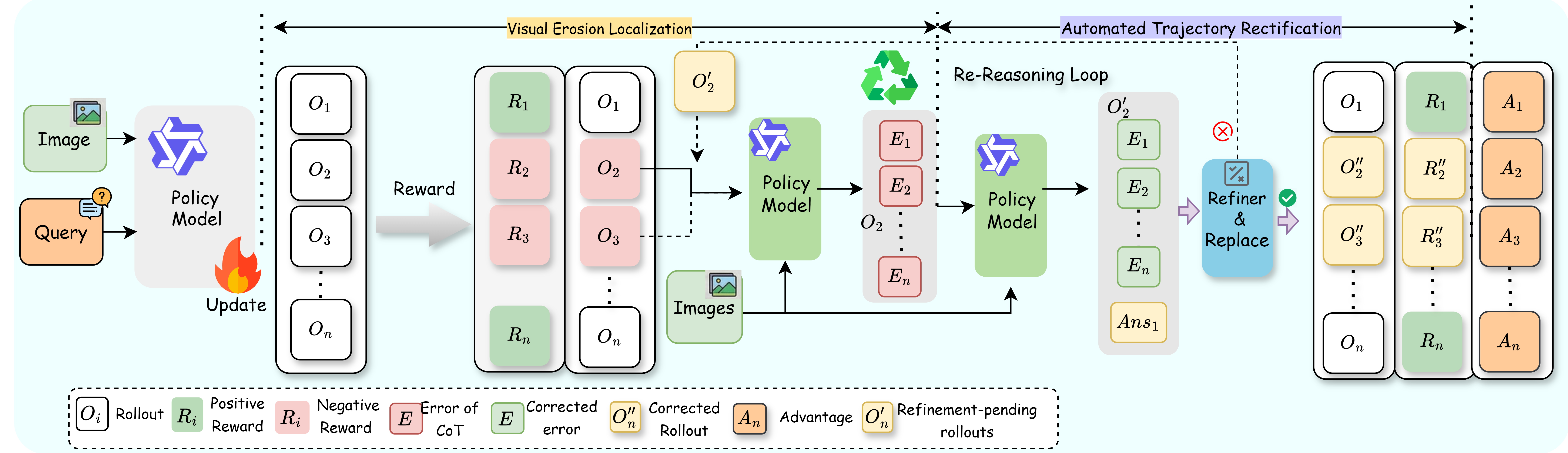} 
  \caption{\textbf{Overview of the IVR-R1 framework.} The pipeline consists of two primary stages: (1) \textbf{Visual Erosion Localization}, which identifies faulty reasoning rollouts ($O_i$) using reward signals; and (2) \textbf{Automated Trajectory Rectification}, where a \textit{Re-Reasoning Loop} iteratively corrects CoT errors ($E_i$) by re-anchoring to the original image, then the \textit{Refiner \& Replace} module synthesizes optimized trajectories ($O''_n$) with calculated advantages ($A_n$) to facilitate iterative policy updates.}
  \label{fig:method} 
  \vspace{-1em}
\end{figure*}

Our contributions are summarized as follows:
\begin{itemize}[leftmargin=*, itemsep=2pt, topsep=2pt, parsep=0pt]
\item We formalize the phenomenon of \textbf{visual grounding erosion} stemming from modality information asymmetry, providing a novel perspective on why MLLMs struggle with logical-visual consistency in long-horizon reasoning.
\item We propose \textbf{visual erosion localization}, which cross-references internal reasoning chains with pristine visual features, thus enabling the model to autonomously perform step-level error attribution and pinpointing exactly where and why the reasoning decoupled from the visual evidence.
\item We propose \textbf{automated trajectory rectification}, which
transforms flawed on-policy rollouts into high-fidelity expert demonstrations, thus offering a scalable, self-bootstrapping alternative to reliance on external black-box models for data purification and trajectory enhancement.
\item Extensive experiments across diverse benchmarks show that IVR-R1 consistently outperforms competing methods, validating the effectiveness of both visual erosion localization and automated trajectory rectification, and establishing a new standard for multimodal RL.
\end{itemize}
\section{Related Work}

The rapid evolution of Multimodal Large Language Models (MLLMs) has transitioned from basic instruction following toward complex, long-horizon reasoning capabilities~\cite{luo2025unlockingmultimodalmathematicalreasoning,NEURIPS2022_11332b6b,bai2025qwen25vltechnicalreport,li2025surveystateartlarge}. Current research addresses this progression through three interconnected pillars: \textbf{(1) post-training alignment} to harmonize model behavior with human intent, \textbf{(2) structured reinforcement learning} to provide granular feedback for intricate tasks, and \textbf{(3) visual representation alignment} to ensure internal reasoning remains anchored to high-fidelity visual evidence. Together, these domains aim to bridge the gap between high-dimensional perception and discrete logical inference, mitigating the "internal state drift" and "grounding erosion" that frequently lead to hallucinations in current multimodal architectures.

\subsection{Post-Training alignment}
Recent advancements in MLLMs have increasingly prioritized post-training alignment techniques, such as instruction tuning~\cite{zhao2024aligngptmultimodallargelanguage} and reinforcement learning (RL), to bolster multimodal performance. While early methods relied on supervised fine-tuning with synthetic data to emulate high-level visual assistant capabilities ~\cite{dai2023instructblipgeneralpurposevisionlanguagemodels,chen2024internvlscalingvisionfoundation}, recent trends have shifted toward RL to enhance complex reasoning~\cite{huang2026visionr1incentivizingreasoningcapability,xu2025llavacotletvisionlanguage,shen2025vlmr1stablegeneralizabler1style}. These strategies often involve sophisticated reward engineering~\cite{li2025selfrewardingvisionlanguagemodelreasoning}, including step-wise supervision of intermediate reasoning chains, the integration of explicit visual grounding into reward functions, and the use of multi-stage curricula that transition from text-only to multimodal tasks~\cite{li2025selfrewardingvisionlanguagemodelreasoning,xia2025visionaryr1mitigatingshortcutsvisual}. Furthermore, preference-based alignment using AI feedback has emerged as a powerful tool for reducing object hallucinations and improving the reliability of model outputs~\cite{vanniekerk2025posttraininglargelanguagemodels,yu2025rlaifvopensourceaifeedback,luu2025enhancingratingbasedreinforcementlearning}.

\subsection{Structured Reinforcement Learning}
The exploration of reinforcement learning within the MLLMs domain has evolved significantly to address the unique challenges of multimodal perception and logical consistency ~\cite{huang2026visionr1incentivizingreasoningcapability,li2025confidenceneedfewshotrl,shen2025vlmr1stablegeneralizabler1style}. Beyond simple preference signals, current research investigates how to provide denser, more informative feedback for complex visual tasks, where sparse rewards often hinder convergence~\cite{peng2025agenticrewardmodelingintegrating,shao2026spuriousrewardsrethinkingtraining,simonds2025rlsrreinforcementlearningself,luo2025thinkingdriftsevidentialgrounding}. A pivotal development in this area is the move toward reasoning decomposition~\cite{li2025selfrewardingvisionlanguagemodelreasoning,xia2025visionaryr1mitigatingshortcutsvisual}, where intricate multimodal queries are deconstructed into simpler, verifiable sub-steps~\cite{li2025recallsanitycheckrole}. This structured approach allows the RL policy to receive more granular feedback throughout the reasoning process, effectively bridging the gap between raw visual perception and final textual answers~\cite{shao2026spuriousrewardsrethinkingtraining,li2025selfrewardingvisionlanguagemodelreasoning}. By integrating these decomposed signals, MLLMs can achieve superior alignment in long-horizon reasoning tasks and improve their ability to generalize across diverse visual environments without relying on exhaustive external human supervision~\cite{li2025selfrewardingvisionlanguagemodelreasoning}.

\subsection{Visual Representation Alignment}
A significant factor contributing to performance degradation in MLLMs is the information asymmetry between high-dimensional visual scenes and discrete textual proxies ~\cite{sun2025mitigatingvisualforgettingtakealong, peng2026deeperthoughtweakeraim,xie2026lavrscenelatentconditioned}. Current research indicates that while visual information is typically pre-encoded into fixed tokens for downstream reasoning~\cite{yoon2025visualrepresentationalignmentmultimodal,xie2026lavrscenelatentconditioned,lin2026vcocloserlookvisual}, the model's internal state often drifts away from the original visual priors as the reasoning chain unfolds ~\cite{li2025selfrewardingvisionlanguagemodelreasoning, xia2025visionaryr1mitigatingshortcutsvisual}. To address this, current research attempts to mitigate the modality gap by explicitly aligning internal representations with pre-trained vision foundation models (VFMs) to retain fine-grained spatial details ~\cite{yoon2025visualrepresentationalignmentmultimodal, zhang2025pixelstokensbytepairencoding}. However, the broader challenge of \textit{visual grounding erosion} remains prevalent, leading to a progressive decoupling of reasoning logic from visual evidence ~\cite{ye2024xvilacrossmodalityalignmentlarge, sun2026thinkinghurtsmitigatingvisual}. This decoupling frequently causes the model to "hallucinate" absent details or pursue misguided reasoning paths that contradict the visual context ~\cite{sun2025mitigatingvisualforgettingtakealong}. Consequently, current research suggests that even with advanced Chain-of-Thought (CoT) capabilities, MLLMs produce erroneous outputs when their linguistic "thinking" is no longer strictly anchored to the visual scene.
\section{Method}

In this section, we begin with a brief introduction to the Group Relative Policy Optimization (GRPO) paradigm and then detail the \textbf{IVR-R1} (Iterative Visual-grounded Reasoning) framework, including visual erosion localization and automated trajectory rectification.

\subsection{Preliminary: GRPO}
To facilitate efficient policy alignment without the computational burden of an external value critic, we adopt the \textbf{GRPO} paradigm. For each multimodal query $Q = \{i, q\}$, where $i$ and $q$ denote the visual and textual inputs respectively, the policy $\pi_{\theta}$ samples a group of $K$ candidate responses $\mathcal{S}_Q = \{s_1, \dots, s_K\}$. 

The \textit{group-relative advantage} $\hat{A}^{grp}$ for the $k$-th response $s_k$ is estimated by standardizing its reward $r(Q, s_k)$ against the intra-group performance:
\begin{equation}
\label{IVR:eq:A}
\hat{A}^{grp}(Q, s_k) = \frac{r(Q, s_k) - \text{mean}(\{r(Q, s_j)\}_{j=1}^K)}{\text{std}(\{r(Q, s_j)\}_{j=1}^K)}.
\end{equation}
This standardization effectively removes question-level biases and stabilizes the policy gradient. The standard GRPO loss $\mathcal{L}_{\text{GRPO}}$ is defined as:
\begin{equation}
\mathcal{L}_{\text{GRPO}}(\theta) = \mathbb{E}_{Q \sim \mathcal{D}} \left[ \sum_{k=1}^{K} \hat{A}^{grp}(Q, s_k) \log \pi_{\theta}(s_k | Q) - \beta \mathbb{D}_{KL}(\pi_{\theta} \| \pi_{\theta_0}) \right],
\end{equation}
where $\pi_{\theta_0}$ is the reference model and $\beta$ is the Kullback–Leibler (KL) divergence coefficient.

\subsection{Iterative Visual-Grounded Reasoning}

To mitigate the pervasive issue of \textit{visual grounding erosion}—where the reasoning chain gradually decouples from visual evidence during long-horizon inference—we propose the \textbf{IVR-R1} framework. As illustrated in Figure~\ref{fig:method}, the methodology transitions from standard policy gradient methods to a dynamic, self-bootstrapping paradigm. The execution flow is structured into three synergistic phases: (1) Visual Erosion Localization, and (2) Automated Trajectory Rectification.

\smallskip
\subsubsection{\textbf{Visual Erosion Localization}}~
\smallskip

\textbf{Step I: Exploration and Screening: Reward-Driven Error Identification} The process begins with an expansive exploration of the reasoning space for each multimodal query $Q = \{i, q\}$. The policy $\pi_{\theta}$ generates a group of $K$ candidate responses:
\begin{equation}
\mathcal{S}_Q = \{s_1, \dots, s_K\}, \quad s_k \sim \pi_{\theta}(\cdot | Q).
\end{equation}
To identify instances of \textit{visual grounding erosion}, each response $s_k$ is evaluated by a pre-computed scalar reward $r(Q, s_k)$, which serves as a diagnostic signal for the model's performance. Specifically, we flag a rollout as "grounding-eroded" if its reward falls below a correctness threshold $\tau$:
\begin{equation}
\text{Flag } s_k \text{ as failed if } r(Q, s_k) < \tau.
\end{equation}
The core objective of this stage is to precisely pinpoint reasoning failures where the model's linguistic "thinking" has bypassed the actual visual evidence. By isolating these failed rollouts, we establish a targeted set of samples for subsequent rectification, ensuring training resources are focused on correcting flawed multimodal logic.

\textbf{Step II: First Visual Interrogation and Visual Erosion Localization.} 

Building on these identified failures, the \textbf{Re-Reasoning Loop} serves as the core diagnostic engine of IVR-R1, designed to bridge the "reasoning-perception gap" by forcing the model to re-interrogate raw visual signals when its linguistic logic drifts.

For each failed response $s_k$, the model is explicitly prompted to perform a "post-mortem" analysis. The core motivation for this step is the empirical evidence that \textbf{original visual features $i$ possess a significant expressive superiority over textual descriptions}, as demonstrated by the widening performance gap in MMMU-Pro benchmarks ($58.1\%$ vs. $54.9\%$). A more comprehensive set of experiments and a fine-grained analysis regarding the impact of these modality-specific features are provided in the ablation study in Section 4.2.

Unlike standard reasoning which may rely on potentially drifted internal tokens or incomplete textual proxies, this step injects the high-fidelity visual pixels back into the context. The model must cross-reference its failed reasoning chain $s_{k, \text{think}}$ against the actual features of $i$ to generate a fine-grained critique $E_k$:
\begin{equation}
E_k = \text{Critique}(i, q, s_{k, \text{think}}, s_{k, \text{ans}}).
\end{equation}
By re-introducing $i$, the framework provides an objective "ground truth" that allows the model to identify precisely where its linguistic logic contradicted the visual evidence, a process essential for resolving "visual erosion" that text alone cannot fix.

\begin{figure}[t] 
  \centering         
  \includegraphics[width=0.75\linewidth]{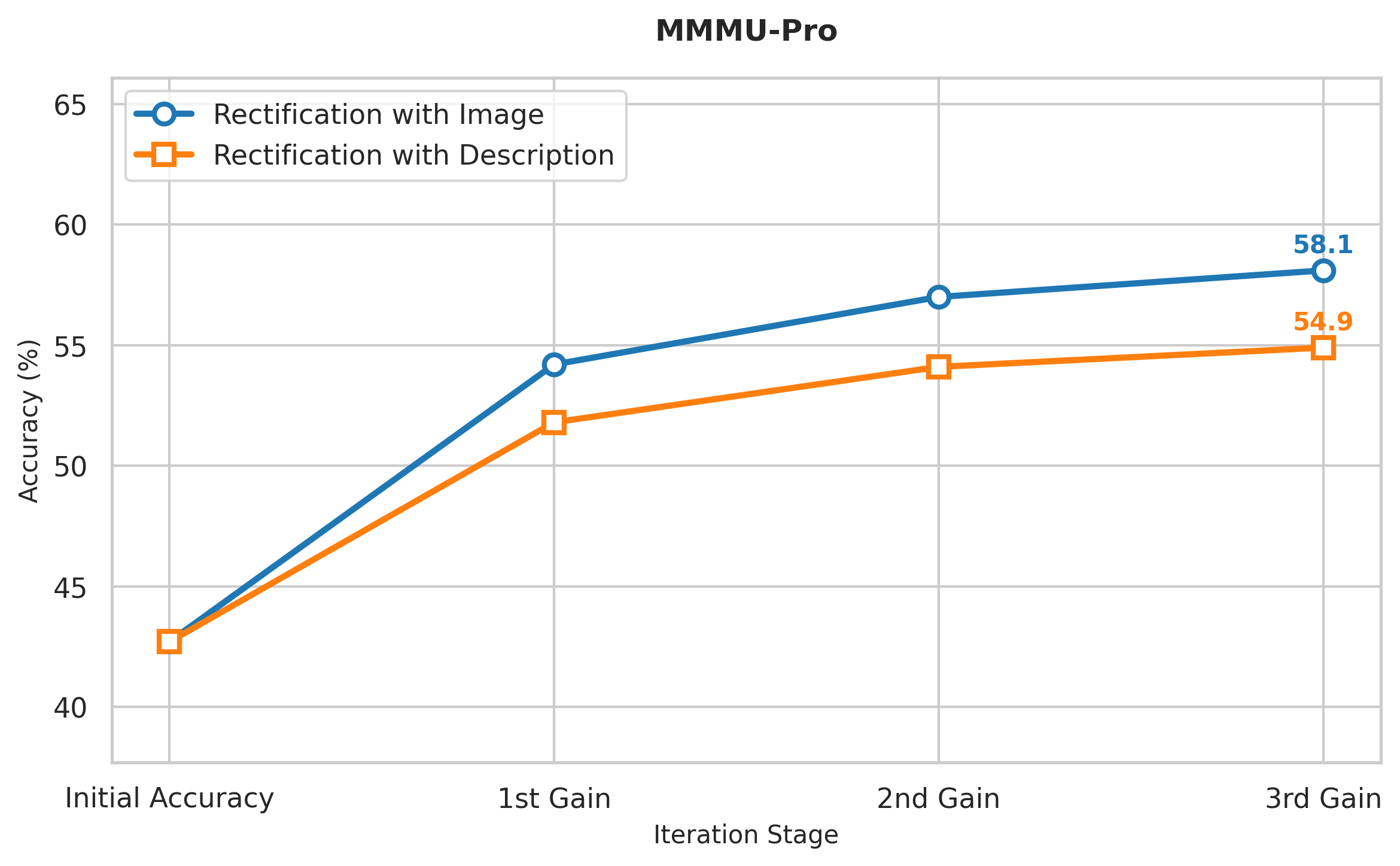} 
  \caption{Ablation study on MMMU-Pro. Image-based rectification ($58.1\%$) significantly outperforms description-based ($54.9\%$) , validating the expressive superiority of raw pixels over textual proxies for trajectory re-anchoring. Detailed experiments and analysis are provided in Section 4.2.}
  \label{fig:data_1} 
\end{figure}

\smallskip
\subsubsection{\textbf{Automated Trajectory Rectification}}~
\smallskip
\textbf{Step I: Secondary Direct Visual Re-Interrogation and Automated Trajectory Rectification}

Guided by the critique $E_k$ and the continuous presence of image $i$, the model reconstructs its reasoning path to produce a rectified response $\hat{s}_k$:
\begin{equation}
\hat{s}_k = \text{Refine}(i, q, E_k, \text{Context}_{ori}).
\end{equation}
In this step, the model is required to re-anchor every logical deduction to specific visual cues identified in $E_k$ by leveraging the context $Context_ori$ of the previously failed rollout $s_k$. 
The purpose is to transform a failed CoT into a \textbf{visually-consistent expert demonstration}. These self-bootstrapped traces $\hat{s}_k$ are logically rigorous and strictly grounded in the visual domain. Because these traces are generated using the model's own internal parameters while being corrected by the objective image $i$, they serve as highly effective training signals.

\textbf{Step II: Policy Shaping via Regularized Optimization}

To internalize the reasoning logic from the rectified traces $\hat{s}_k$, we apply \textbf{Policy Shaping} using a nonlinear transformation $f(x) = \frac{x}{x + \gamma}$ (where $\gamma = 0.1$) on the importance sampling ratio. This transformation increases the gradient emphasis on low-probability yet high-value corrective tokens. The final integrated objective is:
\begin{equation}
\mathcal{J}(\theta) = \mathbb{E}_{Q \sim \mathcal{D}} \left[ \sum_{k=1}^{K} f(\hat{r}_{k}(\theta)) \cdot \hat{A}^{grp}_k - \beta \mathbb{D}_{KL}(\pi_{\theta} \| \pi_{\theta_0}) \right],
\end{equation}
where $\hat{A}^{grp}_k$ is defined in Eq. \eqref{IVR:eq:A}. By re-weighting the policy update, IVR-R1 ensures the model effectively learns unfamiliar but correct grounding decisions, maintaining logical anchoring to visual evidence while following a stable learning trajectory.
\section{Experiments}

In this section, we provide a comprehensive evaluation of the \textbf{IVR-R1} framework. We describe our \textbf{data synthesis strategy}, the \textbf{multi-stage training pipeline}, and the \textbf{diverse benchmarks} spanning mathematical and general multimodal reasoning. Then, we present empirical results and detailed analyses to validate the effectiveness of our approach in mitigating visual grounding erosion and enhancing long-horizon reasoning.

\subsection{Experimental Setup}

\textbf{Model Configurations.} We evaluate our IVR-R1 framework across two representative scales of vision-language models: \textbf{Qwen2.5-VL-7B}~\cite{bai2025qwen25vltechnicalreport} and \textbf{Qwen3-VL-4B}~\cite{bai2025qwen3vltechnicalreport}. These models serve as the policy backbones ($\pi_{\theta}$) due to their robust native multimodal processing capabilities. All experiments are conducted using a distributed training setup to accommodate the iterative generation and optimization phases.

\textbf{Training Pipeline.} Our training follows a two-stage paradigm to ensure both instruction-following stability and reasoning depth:
\begin{enumerate}[leftmargin=*, itemsep=2pt, topsep=2pt, parsep=0pt]
    \item \textbf{SFT Cold-start}: To initialize the model with basic Chain-of-Thought (CoT) reasoning capabilities, we perform a supervised fine-tuning (SFT) warm-up. We utilize a curated set of \textbf{9k high-quality samples} from the \textit{Vision-SR1} dataset~\cite{huang2026visionr1incentivizingreasoningcapability}, which provides structured, step-by-step reasoning trajectories.
    \item \textbf{IVR-R1 RL Optimization}: Following the SFT phase, the model is further optimized using the proposed \textbf{IVR-R1} reinforcement learning framework. This stage leverages the 32k reasoning samples to enhance visual grounding and logical consistency through our re-reasoning loop.
\end{enumerate}

\subsection{Data Composition}

To cultivate robust multimodal logic and alleviate grounding erosion, we curated a diverse and high-quality training corpus for the reinforcement learning phase. We extracted \textbf{32k reasoning-intensive samples} from several state-of-the-art datasets, ensuring a balanced representation of perceptual breadth and logical depth:

\begin{itemize}[leftmargin=*]
    \item \textbf{General Visual Understanding}: We incorporate samples from \textit{LLaVA-CoT-RL}~\cite{xu2025llavacotletvisionlanguage} to provide a foundational multimodal perception. these samples ensure the model maintains broad recognition and descriptive capabilities across a wide range of natural scenes~\cite{xu2025llavacotletvisionlanguage}.
    \item \textbf{Complex Visual Reasoning}: To challenge the model's logical depth, we aggregate samples from \textit{Vision-R1}~\cite{huang2026visionr1incentivizingreasoningcapability}, \textit{MM-K12}~\cite{du2025mmprmenhancingmultimodalmathematical}, and \textit{WeMath}~\cite{qiao2024wemathdoeslargemultimodal}. These datasets target long-horizon reasoning, geometric analysis, and structured mathematical problem-solving~\cite{du2025mmprmenhancingmultimodalmathematical, huang2026visionr1incentivizingreasoningcapability, qiao2024wemathdoeslargemultimodal}, which are the primary scenarios where visual grounding erosion is most likely to occur.
\end{itemize}

\subsection{Evaluation Benchmarks}

We evaluate IVR-R1 on six representative benchmarks, covering both math and general visual understanding, to assess its performance across diverse visual tasks.

\textbf{Mathematical and Numerical Reasoning.} To test the model's ability to interpret and solve complex mathematical problems, we employ:
\begin{itemize}[leftmargin=*, itemsep=2pt, topsep=2pt, parsep=0pt]
    \item \textbf{MathVista}~\cite{lu2024mathvistaevaluatingmathematicalreasoning}: A comprehensive benchmark for mathematical reasoning in visual contexts.
    \item \textbf{MathVerse}~\cite{zhang2024mathversedoesmultimodalllm}: consists of 2.6K diagram-centric problems (e.g., geometry,functions), each rendered in six visual-text variants to disentangle true visual understandingfrom linguistic shortcuts. Evaluation is based on step-by-step Chain-of-Thought (CoT) correctness.
    \item \textbf{MATH-Vision}~\cite{wang2024measuringmultimodalmathematicalreasoning}: presents 3K competition-grade problems across 16 disciplines. We specifically employ the \textbf{testmini} version to evaluate advanced multimodal reasoning capabilities across five difficulty levels.
\end{itemize}

\textbf{General Multimodal Understanding.} To verify the framework's versatility in expert-level knowledge and general visual tasks, we evaluate on:
\begin{itemize}[leftmargin=*, itemsep=2pt, topsep=2pt, parsep=0pt]
    \item \textbf{MMMU}~\cite{yue2024mmmumassivemultidisciplinemultimodal} tests cross-modal reasoning and subject knowledge with 11.5K college-level, four-choice questions spanning six disciplines.
    \item \textbf{MMMU-Pro}~\cite{yue2025mmmuprorobustmultidisciplinemultimodal} increases the difffculty with ten choices per question and adds a challenging vision-only setting, where all text is embedded within the image to necessitate robust visual parsing.
    \item \textbf{MMStar}~\cite{chen2024rightwayevaluatinglarge}: A curated benchmark designed to evaluate "true" multimodal capabilities by filtering out samples solvable through language alone.
\end{itemize} 

\begin{table}[htbp]
  \centering
  \small 
  \setlength{\tabcolsep}{3.5pt}
  \caption{Main comparison of IVR-R1 against baseline methods and open-source models across mathematical reasoning (MathVista, MathVerse, MathVision) and general visual understanding (MMStar, MMMU-Pro, MMMU) benchmarks. To ensure a fair comparison, all reinforcement learning methods are trained on our 32k reasoning dataset. All results are reported under a unified setting of Temperature = 0 with a single sampling trial ($N=1$). Bold values indicate the best performance within each backbone category.}
    \begin{tabular}{cccccccc}
    \toprule
    \multirow{2}{*}{\textbf{Models}} & \multicolumn{3}{c}{\textbf{Math}} & \multicolumn{3}{c}{\textbf{General Visual Understanding}} & \multirow{2}{*}{\textbf{avg}} \\
\cmidrule{2-7}          & MathVista & MathVerse & \multicolumn{1}{c|}{MathVision} & MMStar & MMMU-Pro & MMMU  &  \\
    \midrule
    Qwen2.5-VL-72B-Instruct~\cite{bai2025qwen25vltechnicalreport} & 48.1  & 40.9  & 34.5  & 45.5  & 36.6  & 34.9  & 40.1  \\
    Qwen2.5-VL-32B-Instruct~\cite{bai2025qwen25vltechnicalreport} & 37.2  & 24.3  & 23.4  & 48.3  & 21.7  & 22.9  & 29.6  \\
    Qwen3-VL-32B-Instruct~\cite{bai2025qwen3vltechnicalreport} & 68.7  & 37.4  & 28    & 69.8  & 56.9  & 58    & 53.1  \\
\midrule    Methods & \multicolumn{6}{c}{Backbone model: Qwen2.5-VL-7B			} &  \\
    Zero-shot Inference (before RL) & 32.6  & 17.4  & 21.1  & 41.7  & 16.2  & 21.6  & 25.1  \\
    Supervised Fine-tuning (before RL) & 63.6  & 40.7  & 28.9  & 56.7  & 43.6  & 46.1  & 46.6  \\
    Vision-R1~\cite{huang2026visionr1incentivizingreasoningcapability} &  67.4  &  \textbf{43}  &   30.3    &  58.3  & 47.2 & 50.9  & 49.5  \\
    Vision-SR1~\cite{li2025selfrewardingvisionlanguagemodelreasoning} & 63.5  & 41.9  & 30.6  & 57.7  & 45.5  & 47.6  & 47.8  \\
    \textbf{IVR-R1 (our method)} & \textbf{67.9}\uparrowtext{4.4}  & \textbf{42.6}\uparrowtext{0.7}  & \textbf{31.3}\uparrowtext{0.7}  & \textbf{61.1}\uparrowtext{3.4}  & \textbf{47.8}\uparrowtext{2.3}  & \textbf{53.6}\uparrowtext{5.0}  & \textbf{50.7}\uparrowtext{2.9}  \\
    \midrule
     Methods & \multicolumn{6}{c}{Backbone model: Qwen3-VL-4B} &  \\
    Zero-shot Inference (before RL) & 54.5  & 29.4  & 18.8  & 58.3  & 42.7  & 47.4  & 41.9  \\
    Supervised Fine-tuning (before RL) & 65.6  & 44.4  & 29.3  & 58.9  & 48.6  & 48.4  & 49.2  \\
    Vision-R1~\cite{huang2026visionr1incentivizingreasoningcapability} & 68.3  & 47.5  & 32.6  & \textbf{63.8} & 52.5  & 54.1  & 53.1  \\
    Vision-SR1~\cite{li2025selfrewardingvisionlanguagemodelreasoning} & 68.2  & 46.2  & \textbf{35.5} & 63.1  & 49.3  & 52.4  & 52.5  \\
    \textbf{IVR-R1 (our method)} & \textbf{69.6}\uparrowtext{1.4} & \textbf{49.4}\uparrowtext{3.2} & 35.2\downarrowtext{0.3}  & \textbf{63.3}\uparrowtext{0.2}  & \textbf{53.3}\uparrowtext{4.0} & \textbf{56}\uparrowtext{3.6} & \textbf{54.5}\uparrowtext{2.0} \\
    \bottomrule
    \end{tabular}%
  \label{tab:addlabel}%
\end{table}%

\subsection{Main Results}

\textbf{Overall Performance.} As illustrated in Table~\ref{tab:addlabel}, \textbf{IVR-R1} consistently outperforms both SFT and RL baselines across different model scales. On the \textbf{Qwen2.5-VL-7B} backbone, IVR-R1 achieves a \textbf{50.5\%} average score, marking a \textbf{+3.9\%} absolute improvement over the SFT baseline. Remarkably, \textbf{IVR-R1 (4B)} attains an average of \textbf{54.5\%}, even surpassing the performance of the significantly larger \textbf{Qwen3-VL-32B-Instruct} (53.1\%). This result underscores that high-fidelity reasoning alignment can be more effective than raw parameter scaling in enhancing multimodal intelligence.

\textbf{Domain-Specific Excellence.} The superiority of IVR-R1 is most pronounced in benchmarks requiring tight visual-logic coupling:
\begin{itemize}[itemsep=2pt, topsep=2pt, parsep=0pt]
    \item \textbf{Logical Anchoring:} On \textbf{MathVerse}, IVR-R1 (4B) reaches \textbf{49.4\%}, significantly exceeding the SFT baseline (44.4\%). This demonstrates its resilience against ``internal state drift'' during complex, multi-step reasoning chains.
    \item \textbf{Visual Fidelity:} On \textbf{MMMU}, the 4B model achieves \textbf{56.0\%}, outperforming \textbf{Vision-R1} (+1.9\%) and \textbf{Vision-SR1} (+3.6\%). This indicates a superior ability to preserve and utilize professional-grade visual details during the reasoning process.
\end{itemize}

\textbf{The Re-Reasoning Advantage.} Unlike \textbf{Vision-R1}, which primarily incentivizes reasoning length, IVR-R1 maintains balanced performance across both mathematical and general visual tasks. The core of this success lies in our \textbf{Re-Reasoning Loop}, which effectively mitigates \textbf{visual grounding erosion}. By re-anchoring logical trajectories to the original visual evidence in a self-bootstrapping manner, IVR-R1 achieves superior alignment and generalization without requiring external expert supervision or human-annotated rationales.

\section{Ablation Study}

To further investigate the contribution of each component within the IVR-R1 framework, we perform extensive ablation studies. We first analyze the impact of the number of rectified trajectories, seeking to identify the sweet spot between diversity of expert demonstrations and training stability. Subsequently, we justify the necessity of our visual re-anchoring mechanism by comparing the rectification performance using original visual features against discrete textual proxies. Together, these results provide a comprehensive understanding of how automated trajectory rectification facilitates iterative self-improvement in Multimodal large language models.

\begin{table}[htbp]
  \centering
  \small 
  \setlength{\tabcolsep}{3.5pt}
  \caption{Ablation study on the number of rectified responses in IVR-R1 using the Qwen3-VL-4B backbone. With a sampling group size of $N=8$ and a baseline accuracy of approximately 50\%, we compare the impact of rectifying one, two, and multiple (up to 4) failed trajectories. Bold values indicate the best performance for each benchmark.}
    \begin{tabular}{cccccccc}
    \toprule
    \multirow{2}{*}{\textbf{Pattern}} & \multicolumn{3}{c}{\textbf{Math}} & \multicolumn{3}{c}{\textbf{General Visual Understanding}} & \multirow{2}{*}{avg} \\
\cmidrule{2-7}          & MathVista & MathVerse & \multicolumn{1}{c|}{MathVision} & MMStar & MMMU-Pro & MMMU  &  \\
    \midrule
    One response for IVR & 68.5  & 48.1  & 32.2  & 64.4  & 52.7  & 56.1  & 53.7  \\
    Two responses for IVR & 69.6  & \textbf{49.4} & \textbf{35.2} & 63.3  & \textbf{53.3} & 56    & \textbf{54.5 } \\
    Multiple responses for IVR & \textbf{70.8} & 48.2  & 29.9  & \textbf{64.6} & 52.5  & \textbf{56.5} & 53.8  \\
    \bottomrule
    \end{tabular}%
  \label{tab:number}%
\end{table}%

\subsection{Impact of the Number of Rectified Trajectories}
A key component of the IVR-R1 framework is the transformation of failed rollouts into expert demonstrations. Given that our base model, Qwen3-VL-4B~\cite{bai2025qwen3vltechnicalreport}, exhibits an accuracy of approximately 50\%, a group of $N=8$ samples typically yields about 4 incorrect responses. We investigate the optimal number of such trajectories to rectify within each training step, comparing settings with one, two, and multiple (up to 4) rectified responses. 

As shown in Table~\ref{tab:number}, the performance follows a non-monotonic trend as the number of rectified samples increases. Transitioning from one to two rectified responses results in a significant improvement in the average score (from 53.7\% to 54.5\%), with peak performances in \textbf{MathVerse} (49.4\%) and \textbf{MMMU-Pro} (53.3\%). This suggests that providing a moderate diversity of self-bootstrapped expert signals helps the model internalize robust grounding patterns. 

\begin{figure*}[htbp] 
  \centering         
  \includegraphics[width=1\linewidth]{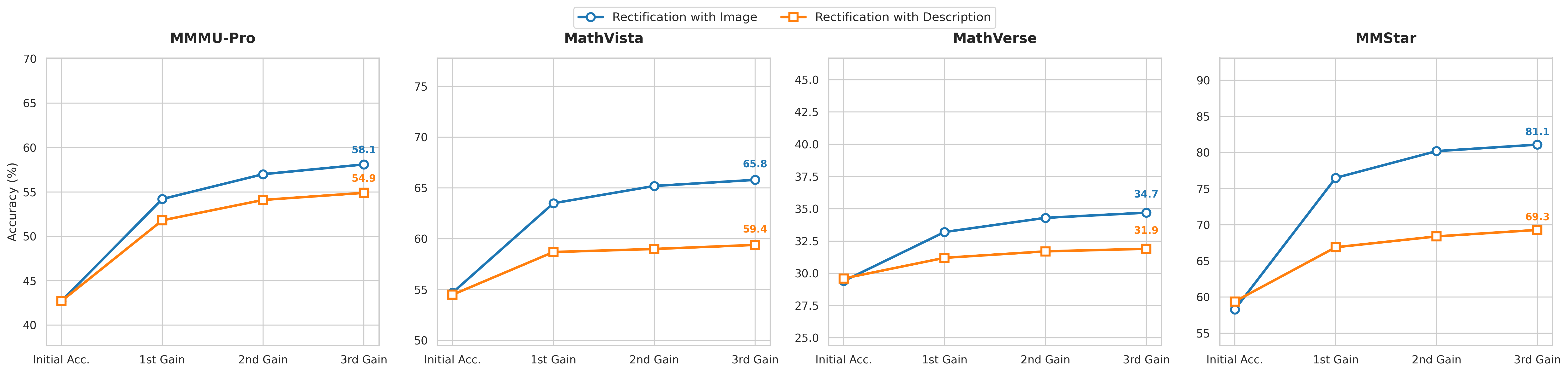} 
  \caption{Ablation study comparing the impact of modality-specific feedback on rectification accuracy across four benchmarks (MMMU-Pro, MathVista, MathVerse, and MMStar). We contrast the performance of the \textbf{Rectification with Image} (blue) against \textbf{Rectification with Description} (orange) under a deterministic setting (Temperature=0, single sampling). The results demonstrate that direct access to original visual features significantly outperforms textual proxies, validating the necessity of visual re-anchoring in our framework.}
  \label{fig:data_2} 
  \vspace{0.3em}
\end{figure*}

However, when increasing the rectification count to "multi" (up to 4), we observe a slight degradation in the overall average (53.8\%), despite gains in specific benchmarks like \textbf{MathVista} (70.8\%) and \textbf{MMMU} (56.5\%). This phenomenon indicates a potential trade-off: while more rectified traces provide more positive signals, they may also introduce higher variance or redundant gradients that slightly destabilize the policy shaping process. Consequently, we select \textbf{two rectified responses} as our default configuration to balance learning efficiency with visual-logic consistency.

\subsection{Superiority of Visual Features in Trajectory Rectification}

To further investigate the role of multimodal information in the \textbf{Automated Trajectory Rectification} process, we conduct a comparative analysis between two input configurations: (1) providing the model with original image pixels (\textit{Rectification with Image}) and (2) providing high-quality textual descriptions of the scene (\textit{Rectification with Description}). This experiment, performed on Qwen3-VL-4B at Temperature=0, aims to quantify the \textbf{information loss associated with discrete textual proxies} and validate our hypothesis regarding the necessity of raw visual anchoring.

As illustrated in Figure~\ref{fig:data_2}, while both configurations show steady accuracy improvements over the \textit{Initial Acc.}, a profound performance gap emerges as the rectification iterations progress. Specifically, the \textbf{Rectification with Image} setting consistently achieves a significantly higher performance ceiling, whereas the description-based approach appears to encounter a \textbf{"semantic bottleneck."} For instance, on the \textbf{MMStar} benchmark—which is specifically curated to evaluate vision-centric tasks that are unsolvable by LLMs alone—utilizing original visual features leads to a peak accuracy of \textbf{81.1\%}, outperforming the description-based approach by a substantial \textbf{11.8\%} margin. Similar trends are observed in \textbf{MathVista} (+6.4\% lead) and \textbf{MMMU-Pro} (+3.2\% lead). These disparities suggest that even the most meticulous textual descriptions suffer from \textbf{"lossy translation,"} failing to capture the intrinsic high-dimensionality and spatial-logical coherence present in raw pixels.

These results yield three critical insights:
\begin{itemize}[leftmargin=*, itemsep=2pt, topsep=2pt, parsep=0pt]
    \item \textbf{Textual Proxies as a Semantic Bottleneck:} The 11.8\% performance gap in MMStar confirms that textual descriptions act as a low-bandwidth proxy. While text can summarize \textit{what} is in an image, it often fails to convey the \textit{spatial-relational} nuances (e.g., precise geometric intersections or subtle texture gradients) required for complex deduction. This empirical evidence supports our claim that information asymmetry between modalities is a primary driver of reasoning failure.
    
    \item \textbf{Re-anchoring to Counteract Internal Drift:} The superior performance of image-based rectification proves that "visual erosion" cannot be mitigated by linguistic logic alone. By re-accessing the original pixels, the model can verify its internal "thinking" against high-fidelity ground truth. This process effectively resets the \textbf{internal state drift}, ensuring that the corrected trajectory is not merely a linguistically plausible guess, but a visually grounded conclusion.
    
    \item \textbf{High-Fidelity Demonstration Synthesis:} The steep trajectory from \textit{Initial Acc.} to \textit{3rd Gain} demonstrates that the IVR-R1 framework effectively facilitates iterative self-improvement. By leveraging original images, the framework transforms failed reasoning into \textbf{high-fidelity expert demonstrations}. These visual-centric signals provide a superior learning signal for policy optimization, which textual descriptions fail to replicate because they lack the raw evidence necessary for the model to "see" why its previous logic was flawed.
\end{itemize}
\section{Conclusion}
In this paper, we present IVR-R1 (Iterative Visual-grounded Reasoning), a reinforcement learning framework designed to mitigate visual grounding erosion in Multimodal Large Language Models. By addressing the information asymmetry between high-dimensional visual inputs and discrete textual reasoning, IVR-R1 ensures that reasoning trajectories remain grounded in visual evidence. Our approach consists of two synergistic stages. Visual Erosion Localization identifies grounding-eroded rollouts via reward-based screening and performs step-level error attribution by re-injecting visual features into the reasoning process. Automated Trajectory Rectification then refines flawed trajectories through a Re-Reasoning Loop, producing high-quality, visually grounded demonstrations without relying on external models.
Experiments on six benchmarks, including MMMU, MathVista, and MathVerse, demonstrate that IVR-R1 achieves state-of-the-art performance in multimodal reinforcement learning, while also revealing that direct visual re-anchoring is essential, as raw visual features provide a substantially higher performance ceiling than textual abstractions. These findings establish IVR-R1 as a robust paradigm for maintaining logical and visual consistency in complex, long-horizon reasoning tasks.
\appendix

\section{Appendix}

\subsection{Composition of the Reward Function}

To encourage the model to generate structured reasoning trajectories while maintaining high task accuracy, we design a composite reward function. The total reward $R_{total}$ for a given prediction $y$ and ground truth $g$ is defined as a weighted sum:

\begin{equation}
R_{total}(y, g) = (1 - \lambda) \cdot R_{acc}(y, g) + \lambda \cdot R_{format}(y)
\end{equation}

where $\lambda$ is a hyper-parameter that balances formatting constraints and task correctness. Following our implementation, we set $\lambda = 0.1$.

\paragraph{Format Reward ($R_{format}$)} 
The format reward ensures that the model strictly adheres to the mandated three-stage output structure: (1) a visual description block, (2) a chain-of-thought reasoning block, and (3) a boxed final answer. Formally, $R_{format}(y)$ is a binary reward:
\begin{equation}
R_{format}(y) = 
\begin{cases} 
1.0 & \text{if } y \text{ matches } \mathcal{P} \\
0.0 & \text{otherwise}
\end{cases}
\end{equation}
where $\mathcal{P}$ denotes the template: \texttt{<desc>}...\texttt{<think>}...\texttt{\textbackslash boxed\{...\}}.
This reward serves as a structural regularizer, forcing the model to explicitly anchor its reasoning in visual observations before proceeding to logic derivation.

\paragraph{Accuracy Reward ($R_{acc}$)} 
The accuracy reward evaluates the correctness of the final conclusion. The system first extracts the content within the \texttt{\textbackslash boxed\{\}} delimiter from the model's output. The extracted answer $\hat{a}$ is then compared against the ground truth $g$ using a task-specific grader:

\begin{equation}
R_{acc}(y, g) = 
\begin{cases} 
1.0 & \text{if } \text{Grade}(\hat{a}, g) \text{ is True} \\
0.0 & \text{otherwise}
\end{cases}
\end{equation}
If no \texttt{\textbackslash boxed\{\}} content is found, the accuracy reward defaults to $0.0$.

\subsection{Experimental Settings and Evaluation Metrics}
\label{appendix:experiment_details}

In this section, we provide additional details regarding the experimental configurations to ensure transparency and reproducibility:

\begin{itemize}
    \item \textbf{Training Epochs:} In all experiments, the Reinforcement Learning (RL) phase is conducted for only \textbf{1 epoch}.
    \item \textbf{Baseline Alignment:} For a fair comparison, all baseline models are trained using the identical \textbf{SFT data} and \textbf{RL data} as our proposed framework.
    \item \textbf{Training Hyperparameters:} During the RL training sampling phase, we set the \texttt{temperature} to $1.0$ and \texttt{top\_p} to $0.99$. For each problem, \textbf{8 responses} are sampled to compute rewards.
    \item \textbf{Evaluation Settings:} To facilitate result reproduction, we use \texttt{temperature: 0.0} for all models during evaluation. Due to limited computational resources, we do not employ an LLM-as-a-judge approach; instead, we utilize the same automated accuracy calculation method used during training.
\end{itemize}


\subsection{Prompts}
\label{appendix:prompts}

This section lists the primary prompts used in our framework. To ensure readability and maintain original formatting, long single-line prompts are automatically wrapped using the \texttt{listings} environment within the styled blocks.

\paragraph{Prompt for Response Generation (Training Phase)}
The following prompt is used during the Reinforcement Learning training phase to guide the model in generating structured responses.

\begin{tcolorbox}[
    promptbox,
    colback=gray!2, 
    colframe=gray!25, 
    title=Training Phase Prompt
]
\begin{lstlisting}
Please solve the following math problem. Your output must strictly follow this specific format: <description>Briefly describe the problem</description> <think>Step-by-step reasoning process</think> The final answer is \boxed{result}. Ensure that all mathematical expressions are properly rendered in LaTeX format and that the logic is coherent throughout the entire reasoning trajectory.
\end{lstlisting}
\end{tcolorbox}

\paragraph{Prompt for Visual Erosion Localization}
This prompt is utilized within the Visual Erosion Localization module to identify and analyze specific regions within the visual embedding space.

\begin{tcolorbox}[
    promptbox,
    colback=blue!2!gray!2, 
    colframe=blue!15!gray!25, 
    title=Visual Erosion Localization Prompt
]
\begin{lstlisting}
As an AI expert, analyze the following incorrect mathematical reasoning based on the provided image.
Question: {Question}
Incorrect Reasoning: {Original_CoT}
Incorrect Answer: {Original_Answer}

Please carefully examine the image and the incorrect reasoning. Identify the exact mistakes made. 
List the errors clearly as:
Error 1: ...
Error 2: ...
\end{lstlisting}
\end{tcolorbox}

\noindent
\begin{minipage}{\linewidth} 
    \paragraph{Prompt for Automated Trajectory Rectification}
    The following prompt guides the model in rectifying reasoning trajectories based on detected inconsistencies or errors.
    
    \vspace{8pt} 
    \begin{tcolorbox}[
        promptbox,
        colback=green!2!gray!2, 
        colframe=green!15!gray!25, 
        title=Automated Trajectory Rectification Prompt,
        before skip=0pt 
    ]
    \begin{lstlisting}
Based on the image, the original question, the incorrect attempt, and the error analysis, please provide the correct step-by-step reasoning and the final answer.
Question: {Question}
Incorrect Attempt: {Original_CoT}
Error Analysis: {Error_Analysis}

Please avoid the mistakes mentioned in the error analysis.
Think step by step and enclose your final answer in \boxed{}.
Output format:
<description> [New visual description] </description> <think> [New and correct reasoning process, without referencing the previous attempt] </think> Final Answer: \boxed{[Answer]}
    \end{lstlisting}
    \end{tcolorbox}
\end{minipage}


\medskip

{\small
 \bibliographystyle{ieee}
 \bibliography{main}
}

\end{document}